\documentclass[letterpaper, 10 pt, conference]{ieeeconf}  % Comment this line out
\IEEEoverridecommandlockouts                              % This command is only
\title{When Shall I Be Empathetic? The Utility of Empathetic Parameter Estimation in Multi-Agent Interactions}

\author{Yi Chen$^{1}$, Lei Zhang$^{1}$, Tanner Merry$^{1}$, Sunny Amatya$^{2}$, Wenlong Zhang$^{2}$, and Yi Ren$^{1}$% <-this % stops a space
\thanks{This work was supported by the National Science Foundation under Grant CMMI-1925403.}
\thanks{$^{1}$Y. Chen, L. Zhang, T. Merry, and Y. Ren are with Department of Mechanical and Aerospace Engineering, Arizona State University, Tempe, AZ, 85287, USA. Email:
        {\tt\small ychen837@asu.edu};
        {\tt\small tmerry@asu.edu};
        {\tt\small lzhan300@asu.edu};
        {\tt\small yiren@asu.edu}}%
\thanks{$^{3}$S. Amatya and W. Zhang are with The Polytechnic School, Ira A. Fulton Schools of Engineering, Arizona State University, Mesa, AZ, 85212, USA. Email:
        {\tt\small samatya@asu.edu};
        {\tt\small wenlong.zhang@asu.edu}}%   
}

\usepackage[utf8]{inputenc}
\usepackage{amssymb}
\usepackage{amsmath}
\usepackage{graphicx}  %Required
\usepackage{xcolor}
\usepackage{hyperref}
\usepackage[noadjust]{cite}
\DeclareMathOperator*{\argmax}{arg\,max}

\usepackage[linesnumbered,ruled]{algorithm2e} 
\usepackage{color}

\iftrue % use space-saving macro
        \newcommand{\cutsectionup}{\vspace*{-0.1in}}
        \newcommand{\cutsectiondown}{\vspace*{-0.07in}}

        \newcommand{\cutsubsectionup}{\vspace*{-0.09in}}
        \newcommand{\cutsubsectiondown}{\vspace*{-0.06in}}

        \newcommand{\cutparagraphup}{\vspace*{-0.17in}}
        \newcommand{\cutparagraphdown}{\vspace*{-0.03in}}

        \newcommand{\cutcaptionup}{\vspace*{-0.1in}}
        \newcommand{\cutcaptiondown}{\vspace*{-0.2in}}

        \newcommand{\cutequationup}{\vspace*{-0.07in}}
        \newcommand{\cutequationdown}{\vspace*{-0.07in}}

        \newcommand{\cuttableup}{}
        \newcommand{\cuttabledown}{}

        \newcommand{\cut}{{\vspace*{-0.02in}}}
        \newcommand{\cutmore}{{\vspace*{-0.06in}}}
        \newcommand{\negcut}{}

\else % do not use space-saving macro
        \newcommand{\cutsectionup}{}
        \newcommand{\cutsectiondown}{}

        \newcommand{\cutsubsectionup}{}
        \newcommand{\cutsubsectiondown}{}

        \newcommand{\cutparagraphup}{
        \newcommand{\cutparagraphdown}{}

        \newcommand{\cutcaptionup}{}
        \newcommand{\cutcaptiondown}{}

        \newcommand{\cutequationup}{}
        \newcommand{\cutequationdown}{}

        \newcommand{\cuttableup}{}
        \newcommand{\cuttabledown}{}

        \newcommand{\cut}{}
        \newcommand{\cutmore}{}
        \newcommand{\negcut}{}
\fi

\begin{document}

\maketitle
\thispagestyle{empty}
\pagestyle{empty}

\begin{abstract}
Human-robot interactions (HRI) can be modeled as dynamic or differential games with incomplete information, where each agent holds private reward parameters. Due to the open challenge in finding perfect Bayesian equilibria of such games, existing studies often consider approximated solutions composed of parameter estimation and motion planning steps, in order to decouple the belief and physical dynamics. In parameter estimation, current approaches often assume that the reward parameters of the robot are known by the humans. We argue that by falsely conditioning on this assumption, the robot performs non-empathetic estimation of the humans' parameters, leading to undesirable values even in the simplest interactions. We test this argument by studying a two-vehicle uncontrolled intersection case with short reaction time. Results show that when both agents are unknowingly aggressive (or non-aggressive), empathy leads to more effective parameter estimation and higher reward values, suggesting that empathy is necessary when the true parameters of agents mismatch with their common belief. The proposed estimation and planning algorithms are therefore more robust than the existing approaches, by fully acknowledging the nature of information asymmetry in HRI. Lastly, we introduce value approximation techniques for real-time execution of the proposed algorithms.
\end{abstract}

% \cutsectionup
\section{Introduction}
\label{sec:intro}
%\cutsectiondown
Human-robot interactions (HRI) have become ubiquitous in the past two decades, with applications in daily assistance, healthcare, manufacturing, and defense. Since humans and robots may not understand intents and strategies of each other when completing collaborative tasks, we consider HRI as dynamic general-sum games with incomplete information, where agents hold private parameters that determine their rewards. Finding perfect Bayesian equilibria (PBE) of such games is an open challenge~\cite{buckdahn2011some} due to the entanglement of physical and belief dynamics, and existing solutions (e.g., structured PBEs) do not scale well with the dimensionalities of the state, action, and reward parameter spaces~\cite{sinha2016structured,vasal2018systematic}.
% \footnote{There is limited scalability of the fixed point problems underlying the PBEs for which the solutions are entangled measures defined on policy and belief spaces~\cite{***}.}.
As a result, most existing HRI studies either resort to simplified  complete-information games~\cite{foerster_learning_2017,sadigh_planning_2018,kwon2020humans,schwarting2019social}, or adopt variants of an empirical solution composed of two steps, namely, \textit{parameter estimation} and \textit{motion planning}~\cite{nikolaidis2017human,sun2018probabilistic,peng2019bayesian,fridovich2020confidence}. This paper focuses on the latter approach. In the parameter estimation step, each agent takes in past observations of the states and actions of all agents, updates their \textit{common belief} about all agents' parameters, and each derives an estimate (and its uncertainty) of the parameters of their fellow agents. This estimate is then used by the agent to plan its actions.

A common approach to parameter estimation based on noisy (bounded) rationality~\cite{fridovich2020confidence} assumes a Boltzmann distribution on the action space, where the probability of taking any action is determined by the action-value (Q-value), i.e., the equilibrial cumulative reward if the action is taken. Since action-values are parameterized by agents' reward parameters, the belief about these parameters can be computed through Bayes updates using the observed actions (Sec.~\ref{sec:methods}). 
% This approach has been adopted for inverse optimal control~\cite{***} and inverse dynamic games~\cite{***}.

\begin{figure}
    \centering
    \includegraphics[width=\linewidth]{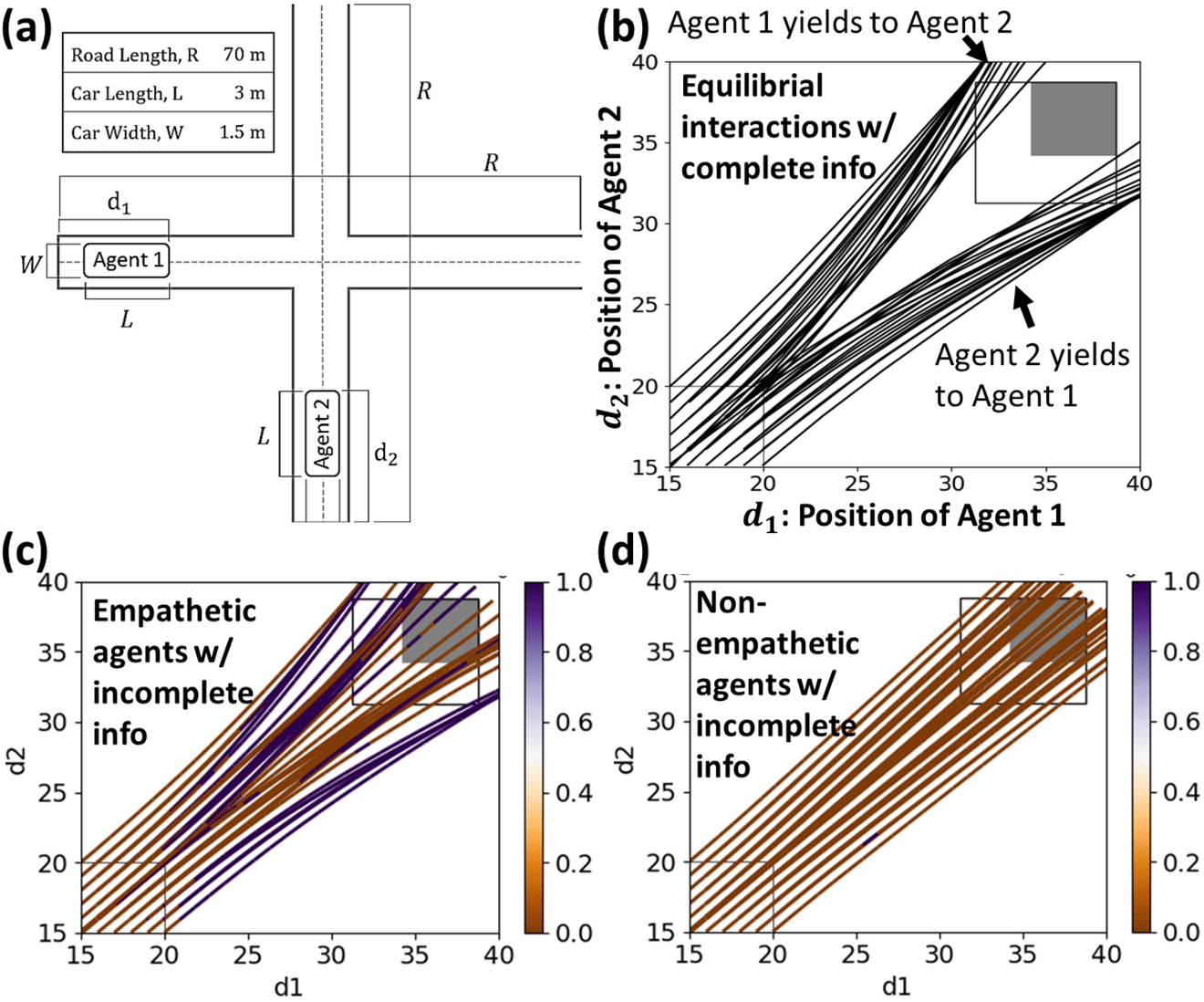}
    \vspace{-0.2in}
    \caption{(a) Schematics of a two-agent uncontrolled intersection case. $d_1$ and $d_2$ are positions of agents. (b) Nash equilibrial state trajectories when both agents know they are non-aggressive. Empathetic (c) and non-empathetic (d) non-aggressive agents when they have incorrect initial common beliefs that both are aggressive (less sensitive to close distances). Empathetic agents are more likely to avoid collisions due to their better estimation of others' reward parameters and choice of the correct policies (color coded as 1-purple). Solid and hollow boxes at the top right corner represent ``close-distance'' states from the perspectives of the aggressive and non-aggressive agents, respectively. Bottom left hollow box represents the initial states. Trajectories are mirrored along the diagonal line due to intrinsic symmetry.}
    \label{fig:summary}
    \vspace{-0.25in}
\end{figure}

Specific to HRI, existing implementations of parameter estimation fall into two categories: In the first, agents assume that their ego parameters are known to their fellows, and estimate fellows' parameters conditioned on this assumption~\cite{nikolaidis2017human,sun2018probabilistic,peng2019bayesian}. This approach set agents to be \textbf{\textit{non-empathetic}}, since they do not consider the potential mismatch between their true parameters and the estimates by their fellows. The second approach removes the conditioning~\cite{liu_who_2016,ren2019icra}. Since the common belief is not necessarily consistent with the true parameters, agents become \textbf{\textit{empathetic}}, i.e., allowing their fellows to have wrong estimates about their parameters. It should be noted that empathy incurs higher computational loads, since it requires full knowledge of the probabilities over the joint parameter space of all agents. In comparison, non-empathetic estimation only requires partial knowledge due to the conditioning.  
% and take actions following the equilibrial Q-values, where Nash equilibrium is often used as a solution concept~\cite{***}. 
% Both settings assume complete-information games, and are thus theoretically incorrect when applied to incomplete-information ones. On the other hand, both have been pragmatic choices for real-time HRI when computational cost is of concern.  
% A significant volume of existing HRI studies resort to non-empathetic models~\cite{***} or even assume complete information~\cite{***}. 
The necessity of empathy was only recently demonstrated for hand-picked interaction settings (e.g., vehicle interactions with specific initial states~\cite{liu_who_2016,ren2019tiv}), which motivates the central question of this paper: 

\begin{center}{\parbox{0.9\linewidth}{\textbf{{\textit{In what interactions does empathy of agents matter?}}}}}\end{center}

%this paper answers a practical yet less discussed question: \textit{Given a set of multi-agent interactions parameterized by agent parameters and initial states, does there exist a subset of interactions where the game-theoretic and non-game-theoretic parameter estimation modules lead to significantly different interactions?} 

This paper makes the following contributions towards answering this question (summary in Fig.~\ref{fig:summary}).
(1) \textbf{Methodology:} we define an interaction space spanned by the initial system states, agent parameters, the common belief about the parameters, and the empathy of agents to systematically evaluate the potential advantages (or disadvantages) of empathy. (2) \textbf{Knowledge:} through a two-vehicle uncontrolled intersection case, we show that empathy in parameter estimation leads to significantly better values when reward parameters and common beliefs are completely wrong (e.g., everyone being aggressive while believing all to be nonaggressive). (3) \textbf{Computation:} to enable fast parameter estimation and motion planning, we develop a learning architecture for approximating Nash equilibrial action-values for given agent parameters. The approximation model is trained on equilibrial interactions solved from the boundary-value problems (BVPs) following Pontryagin's Maximum Principle (PMP)~\cite{pontryagin1966theory} on a meshgrid of the state space. 
% \begin{enumerate}
%     \item \textbf{Methodology:} We define an interactions space spanned by the initial system states, agent parameters, the common belief about the parameters, and the empathy of agents. We propose to use social value, i.e., the total value of all agents, for evaluating the potential advantages (or disadvantages) of empathetic estimation.
    
%     \item \textbf{Knowledge:} Through a two-agent uncontrolled intersection case, we identify regions in the interaction space where empathy leads to higher social value. In particular, we show that (1) empathy leads to interactions with social values no worse than those lack of it; and (2) empathy achieves better social values when agents are (unknowingly) nonaggressive.

%     \item \textbf{Computation:} To enable computationally tractable parameter estimation and motion planning in the context of differential games, we develop a neural network for approximating Nash equilibrial action-values for given agent parameters. The network is trained on a set of equilibrial value trajectories derived by solving the boundary-value problems following Pontryagin's Maximum Principle~\cite{pontryagin1966theory} on a meshgrid of the state space. 
% \end{enumerate}

The rest of the paper is structured as follows: Sec.~\ref{sec:related} reviews related work. Sec.~\ref{sec:methods} elaborates on empathetic and non-empathetic parameter estimation, motion planning, and data-driven action-value approximation. Sec.~\ref{sec:case} introduces the case, experimental settings, hypotheses, and analyses. We conclude the paper in Sec.~\ref{sec:conclusion}. 

\section{Related Work}
\label{sec:related}

\textbf{Multi-agent perfect Bayesian equilibrium:} A PBE consists of a policy and belief pair that simultaneously satisfies sequential rationality and belief consistency~\cite{fudenberg1991perfect}. It is known that there does not exist a universal algorithm for computing PBE due to the interdependence of policies and beliefs~\cite{vasal2018systematic}. This open challenge is partially addressed recently in \cite{vasal2018systematic}, which shows that a subset of PBEs can be computed recursively by solving fixed-point equations for each agent where the solutions are probability measures (rather than point values) conditioned on the agents' private information. Since the fixed-point equations are interdependent on agents' policies, the algorithm is non-scalable with respect to the number of agents, time, or the dimensionalities of the action, state, and parameter spaces. Only solutions for two-agent two-step games have been demonstrated so far~\cite{vasal2018systematic,sinha2016structured}. The inverse problem, i.e., estimation of agent parameters given PBE demonstrations, has not been studied. 

\textbf{Decision modeling:}
Human decision models in HRI~\cite{antonini2006discrete,gupta2018social,bobu2020less} follow studies in choice modeling~\cite{mcfadden2000mixed} and behavioral economics~\cite{su2008bounded,bischi2000global}. Risk models are introduced to explain seemingly non-optimal human actions~\cite{kwon2020humans}. Social value orientation is introduced in \cite{schwarting2019social} to explain agents' courtesy towards others in general-sum dynamic games. Similar courtesy models have been discussed in \cite{sun2018courteous,ren2019tiv}. In this paper we model agents to take Nash equilibrial actions deterministically without considering courtesy. We only use noisy rationality to compensate for modeling errors during belief updates, similar to \cite{fridovich2020confidence}.

% \textbf{Human decision modeling:} A variety of agent parameters have been introduced to explain real-world agent behavior. These include preference weights on individual reward components~\cite{***}, rationality parameter governing noisiness in actions~\cite{***}, risk parameters following behavioral economics~\cite{***}, and social value orientation that weighs ego and fellow values~\cite{***}. 

\textbf{Multi-agent inverse reinforcement learning (MIRL):} Parameter estimation (for reward and policy) has been investigated for single-agent problems~\cite{abbeel2004apprenticeship,ziebart2008maximum,fu2017learning}, under the term of inverse reinforcement learning (IRL). For dynamic games, IRL falsely assumes that each agent uses optimal control while all of its fellow agents follow fixed trajectories. To address this, MIRL performs estimation under solution concepts of the game rather than assuming optimality of individual actions~\cite{lin2014multi,lin2019multi}. Along the same vien, the belief update algorithm introduced in this paper extends the single-agent framework in \cite{fridovich2020confidence} to games, while allowing noisiness of rationality to be estimated along with the agents' reward parameters. Compared with \cite{schwarting2019social}, where agents' parameters are estimated using Stackelberg equilibrial as a solution concept, this paper considers agents to take simultaneous actions and are thus Nash equilibrial.
% As a further comparison with \cite{schwarting2019social}, we also extend our study beyond parameter estimation, to examine the ramifications of empathetic and non-empathetic estimations to a variety of interactions. 

\textbf{Value approximation:} Solutions to Hamilton-Jacobi equations often have no analytical forms, can be discontinuous, and only exist in a viscosity sense~\cite{evans1984differential,lions1985differential}. Deep neural networks (DNN) have recently been shown to be effective at approximating solutions to Hamilton-Jacobi-Bellman (HJB) equations underlying optimal control problems~\cite{nakamura2019adaptive} and Hamilton-Jacobi-Isaac (HJI) equations for two-player zerosum games with complete information~\cite{rubies2016recursive}, thanks to the universal approximation capability of DNNs~\cite{lu2017expressive}. In this paper, we extend this approximation scheme to values of general-sum complete-information differential games, and then use the resultant value networks to approximate agents' action-values during parameter estimation and motion planning. In comparison, \cite{schwarting2019social} requires equilibria to be computed by iteratively solving the KKT problems during the estimation of agent parameters, while the proposed value approximation technique allows agents to memorize values offline, thus accelerating the online estimation. 

% \cutsectionup
\section{Methods}
\label{sec:methods}
This section introduces the parameter estimation and motion planning algorithms to be used in the case study. We also elaborate on the approximation techniques of action-values.

\cutsubsectionup
\subsection{Parameter estimation and motion planning}

%Yi's note
% An agent using noisy-rational decision model would act according to the action with the highest probability.

% \textbf{Motion prediction}: Let the probability of the agent being in state $x(k)$ at time step $k$ be $p(x(k);\beta)$. The probability $p(x(k+1);\beta)$ can be computed using the Kolmogorov forward equation:
% \begin{equation}
% \begin{aligned}
%     p(x(k+1);\beta) = \int_{\mathcal{U}} p(u|x(k);\beta) p(x(k);\beta),
%     \label{eq:motionprediction}
% \end{aligned}
% \end{equation}
% where $x(k)$ is calculated based on $x(k+1)$ and $u$ through the dynamics $h$. The marginal will be
% \begin{equation}
%     p(x(k+1); \mathcal{D}) = \sum_{\beta \in \mathcal{B}} p(x(k+1);\beta) p(\beta; \mathcal{D}).
% \end{equation}

% \subsection{Game-theoretic human-machine interaction model}

\textbf{Preliminaries and notations:} For generality, we consider a multi-agent game with $N$ agents. All agents share the same individual action set $\mathcal{U}$, state space $\mathcal{X}$, reward parameter set $\Theta$, and rationality set $\Lambda$. Together, they share an instantaneous reward function $f(\cdot,\cdot;{\theta}): \mathcal{X}^N \times \mathcal{U}^N \rightarrow \mathbb{R}^N$, a terminal reward function $c(\cdot;{\theta}): \mathcal{X}^N \rightarrow \mathbb{R}^N$, a dynamical model $h: \mathcal{X}^N \times \mathcal{U}^N \rightarrow \mathcal{X}^N$, and a finite time horizon $[0,T]$. Let $\beta_i := <\lambda_i,\theta_i>$ be the parameters of agent $i$, where $\lambda_i \in \Lambda$ and $\theta_i \in \Theta$. We denote the total parameter set by $\mathcal{B}:=\Lambda^N \times \Theta^N$. $\Theta$, $\Lambda$, $\mathcal{B}$, and $\mathcal{U}$ are considered discrete in this study. 
% Agents are defined by $<\beta_i, f_{\theta_i}, c_{\theta_i}>$ for $i=1,...,N$. 
% $\beta:=(\beta_1,...,\beta_N)$, $x := (x_1,..., x_N)$, and $u:=(u_1,...,u_N)$ be , respectively. 
To reduce notational burden, we use a single variable $a$ for the set $(a_1,...,a_N)$ and define $a_{-i}=(a_1,...,a_{i-1},a_{i+1},...a_{N})$. E.g., $\beta \in \mathcal{B}$ contains parameters for all agents, $\beta_{-i}$ those except for agent $i$. We denote by $a^*$ the true value of variable $a$, and $\hat{a}$ its point estimate.
Lastly, we assume the existence of a prior common belief $p_0(\beta)$, which will be updated as $p_k(\beta):= p(\beta;\mathcal{D}(k))$ at time step $k$ with observations $\mathcal{D}(k)=\{(x(t), u(t))\}_{t=1}^k$.

% At time step $k$, agent $i$'s estimation of $\beta_{-i}$ follows 
% \begin{equation}
%     \hat{\beta}_{-i} = \argmax_{\beta_{-i}} p_k(\beta_{-i}|\beta_i^*),
% \end{equation}
% where $\beta_i^*$ are the true parameters of agent $i$. % We assume that $P_H$ is a dirac function on $\Lambda \times \Theta$ that represents a \textit{fixed} belief of the machine, i.e., $P(\hat{\lambda}_M, \hat{\theta}_M) = 1$, while $P_M$ can change along time during an interaction with $H$. This modeling assumption considers the worst-case scenario where the human agent do not update its understanding of the machine's parameters. 

\textbf{Nash equilibria for complete-information game}: If $\theta$ is known to all and unique Nash equilibria exist, agents can derive the equilibrial action-values $Q(\cdot,\cdot;\theta): \mathcal{X}^N \times \mathcal{U}^N \rightarrow \mathbb{R}^N$. For agent $i$, $Q_i(x,u;\theta)$ is the value of action $u_i$ in state $x$, when ego and fellow parameters are $\theta_i$ and $\theta_{-i}$, respectively. For the discrete set of joint parameters, $\Theta^N$, we can derive $\mathcal{Q}^N:=\{Q(\cdot,\cdot;\theta)\}_{\theta \in \Theta^N}$, which maps $\Theta^N$ to the coupled equilibrial action-values. E.g., for a two-agent game where $|\Theta|=2$, we have $|\mathcal{Q}^2|=4$ pairs of action-values. We discuss the approximation of action-values in Sec.~\ref{sec:valueapp}. 

% To facilitate probabilistic inference, the agent is modeled to take action $u \in \mathcal{U}$ under state $x \in \mathcal{X}$ with probability density
% \begin{equation}
%     % \hat{\theta}, \hat{\lambda} = \argmax_{\theta \in \Theta, \lambda \in \Lambda} \frac{\exp(\lambda Q(x,u,\theta))}{\sum_{u \in \mathcal{U}} \exp(\lambda Q(x,u,\theta))} 
%     p(u|x;\beta) = \frac{\exp(\lambda Q(x,u;\theta))}{\int_{\mathcal{U}} \exp(\lambda Q(x,u';\theta))du'},
%     \label{eq:policy}
% \end{equation}

% \begin{equation}
%     p(u|x;\hat{\beta}) = \prod_{i=1,...,N} p(u_i|x;\hat{\beta}).
% \end{equation}

\textbf{Belief update}: Given observations $\mathcal{D}(k)$ at time step $k$, $p_k(\beta)$ follows Bayes update:
\begin{equation}
    p_k(\beta) = \frac{p(u(k)|x(k);\beta)p_{k-1}(\beta)}{\sum_{\beta' \in \mathcal{B}}p(u(k)|x(k);\beta')p_{k-1}(\beta')},
    \label{eq:beliefupdate}
\end{equation}
% Since each $(\beta_H, \hat{\beta}_M)$ may map to multiple Q function pairs in $\mathcal{Q}^2$, we first compute $P(Q_H,Q_M|D(k))$ and then $P(\beta_H, \hat{\beta}_M|D(k))$. Specifically,
% \begin{equation}
%     P(Q_H,Q_M|D(k)) = \frac{P(u_H(k)|x(k);Q_H)P(Q_H,Q_M|D(k-1))}{\sum_{(Q_M', Q_H') \in \mathcal{Q}^2}P(u_H(k)|x(k);Q_H')P(Q_H',Q_M'|D(k-1))},
%     \label{eq:updateQ}
% \end{equation}
% and
% \begin{equation}
% \begin{aligned}
%     P(\beta_H, \hat{\beta}_M|D(k)) = & \sum_{(Q_M', Q_H') \in \mathcal{Q}^2}P(\beta_H, \hat{\beta}_M|Q_H')P(Q_H'|D(k)),
%     \label{eq:updatebeta}
% \end{aligned}
% \end{equation}
% where
% \begin{equation}
%     P(\beta_H, \hat{\beta}_M|Q_H') = \frac{P(Q_H'|\hat{\beta}_M,\beta_H)P(\beta_H, \hat{\beta}_M|D(k-1))}{\sum_{(\beta_H',\hat{\beta}_M) \in \Lambda^2 \times \Theta^2} P(Q_H'| \hat{\beta}_M,\beta_H')P(\beta_H', \hat{\beta}_M|D(k-1))},
% \end{equation}
% %Yi's note
% here, $P(Q_H'|\hat{\beta}_M,\beta_H)$ is related to the Nash Equilibrium; we can have some number of Q pairs in the Nash Equilibrium, and each one of them will share the same amount of probability. \\
% To avoid invalid Bayesian updates of $P(Q_H|D(k))$ and $P(\beta_H|D(k))$ due to observation noises and model discrepancy, we can preprocess these two following Eq.~\eqref{eq:preprocess} before calling Eq.~\eqref{eq:updateQ} and Eq.~\eqref{eq:updatebeta}.
where $p(u|x;\beta) = \prod_{i=1,...,N} p(u_i|x;\beta)$ since actions are modeled to be drawn independently by agents, and
\cutequationup
\begin{equation}
    p(u_i|x;\beta) = \frac{\exp(\lambda_i Q_i(x,(u_i, u^{\dagger}_{-i});\theta))}{\sum_{\mathcal{U}} \exp(\lambda_i Q_i(x,(u_i', u^{\dagger}_{-i});\theta))}.
    % \label{eq:decision}
\cutequationdown
\end{equation}
$\forall i = 1, ..., N$. 
% Here $\bar{Q}_i(x,u_i;\theta)= \left[Q_i(x,u;\theta)\right]$, and $p_{\mathcal{U}^{N-1}}$ is the mixed strategy of fellows of agent $i$. We solve the mixed strategies using Q tables populated by the value approximation network. 
Here $u^{\dagger}_{-i}$ are the observed actions taken by all agents except $i$ at the previous time step. 
Notice that this update is shared among all agents, provided that they receive the same observations. Therefore $p_k(\beta)$ remains a \textit{common} belief. It should also be noted that if $\beta \in \mathcal{B}$ is mistakenly assigned zero probability, e.g., due to noisy observations, this mistake will not be fixed by future updates. To address this, we modify $p_k(\beta)$ as
\cutequationup
\begin{equation}
    p_k(\beta) = (1-\epsilon)p_k(\beta) + \epsilon p_0(\beta)
    \label{eq:preprocess}
\cutequationdown
\end{equation}
before its next Bayes update, where $\epsilon$ represents the learning rate. This allows all $\beta$ combinations to have non-zero probabilities throughout the interaction, provided that the prior $p_0$ is non-zero on $\mathcal{B}$. 

% \textbf{Motion prediction}: Based on $P(\beta_H, \hat{\beta}_M|D(k))$, we can compute the probability of $H$ and $M$ being in state $x(k)$ from $H$'s perspective. First 
% \begin{equation}
% \begin{aligned}
%     P(x(k+1)|Q_H,Q_M) = \sum_{x(k)\in \mathcal{X}, u(k) \in \mathcal{U}} & P(x(k+1)|x(k), u(k)) \\ & P(u_H(k), u_M(k)|x(k);Q_H,Q_M) \\ %changed by Yi: from ; to |
%     & P(x(k)|Q_H,Q_M),
%     \label{eq:motionprediction_game}
% \end{aligned}
% \end{equation}
% where 
% \begin{equation}
%     P(u_H(k), u_M(k)|x(k);Q_H,Q_M) = P(u_H(k)|x(k);Q_H)P(u_M(k)|x(k);Q_M),    
% \end{equation}
% and $(Q_H, Q_M) \in \mathcal{Q}^2$. Then
% \begin{equation}
%     P(x(k+1)|D(k)) = \sum_{(Q_H,Q_M) \in  \mathcal{Q}^2} P(x(k+1)|Q_H,Q_M)P(Q_H,Q_M|D(k)).
% \end{equation}

\textbf{Parameter estimation}: The point estimate
\cutequationup
\begin{equation}
    \hat{\beta}(k) = \argmax_{\beta \in \mathcal{B}} p_k(\beta)
    \label{eq:empathetic}
\cutequationdown
\end{equation}
is empathetic for agent $i$ in the sense that it allows others' estimation of $\beta_i$, which is the $i$th element of $\hat{\beta}(k)$, to be different from the true parameters $\beta_i^*$. Notice that $\hat{\beta}(k)$ is shared among all empathetic agents. For a non-empathetic agent $i$, its own estimation of fellow agents follows
\cutequationup
\begin{equation}
    \tilde{\beta}_{-i}(k) = \argmax_{\beta_{-i} \in \mathcal{B}_{-i}} p_k(\beta_{-i}|\beta_i^*).
    \label{eq:nonempathetic}
\cutequationdown
\end{equation}
Different from empathetic agents, non-empathetic agents may have estimates different from each other, due to the conditioning on their own parameters. 

\textbf{Motion planning}: If agents take actions strictly following the common belief, the interactions will be solely determined by the prior $p_0(\beta)$ independent of the private parameters of agents. This is inconsistent with real-world interactions where agents express their own intents. Therefore, we model each agent $i$ to follow control policies parameterized by their own parameters and the estimates of others, i.e., $\hat{\theta} = (\theta^*_{i}, \hat{\theta}_{-i})$ for the empathetic case and $\hat{\theta} = (\theta^*_{i}, \tilde{\theta}_{-i})$ for the non-empathetic case.
% Note that empathetic agents behave according to their own parameters, even when they permit others' misunderstanding of them.
Specifically, agent $i$ takes actions deterministically following
\cutequationup
\begin{equation}
    u_i = \argmax_{u_i \in \mathcal{U}} Q_i(x,(u_i, u^{\dagger}_{-i});\hat{\theta}).
    \label{eq:decision}
\cutequationdown
\end{equation} 

\textbf{Simulated interactions}: Alg.~\ref{alg:sim} summarizes the simulation of an interaction, which is parameterized by the initial states $x_0$, the set of agent parameters $\beta^*$, and the prior belief $p_0(\beta)$. The simulation outputs the trajectories of states $x(k)$, actions $u(k)$, beliefs $p_k(\beta)$, and values $v(k)$ of all agents. 

\begin{algorithm}[h]
\SetAlgoLined
\SetKwInOut{Input}{input}
\SetKwInOut{Output}{output}
\Input{$x_0$, $\beta^*$, $p_0(\beta)$}
\Output{$\{(x(k), u(k), p_k(\beta), v(k)\}_{k=1}^T$}
    set $k=0$ and $x(0)=x_0$\;
    \While{$k \leq T$}{
        compute $\hat{\beta}_{-i}$ if $i$ is empathetic (or $\tilde{\beta}_{-i}$ if non-empathetic) using Eq.~\eqref{eq:empathetic} or Eq.~\eqref{eq:nonempathetic}\;
        
        compute $u_i(k)$ from Eq.~\eqref{eq:decision}\;
        
        compute $x(k+1) = h(x(k), u(k))$ \;
        
        update $p_{k+1}(\beta)$ using Eq.~\eqref{eq:preprocess} and Eq.~\eqref{eq:beliefupdate} \;
        
        $k = k+1$ \;
    }
\caption{Multi-agent interaction}
\label{alg:sim}
\end{algorithm}
\vspace{-0.1in}
\cutsubsectionup
\subsection{Action-value approximation}
\label{sec:valueapp}
Theoretically, the action-values of incomplete-information dynamic games should be defined on the joint space of physical and belief states. As a simplification, we approximate the action-values using those of the complete-information version of the games, parameterized by the estimated agent parameters. Further, while we model the interactions as (discrete-time) dynamic games, we propose to derive the equilibria from the (continuous-time) differential games to fully exploit the known physical dynamics. In the following, we describe the approaches to solving the resultant BVPs and to learning a value approximation based on the BVP solutions. We will highlight case-specific challenges and our solutions after we introduce the uncontrolled intersection case in Sec.~\ref{sec:case}.
% from \cite{Noncooperative Differential Games} to first derive the equilibrial values, and then approximate the Q-values at discrete time steps based on the equilibrial values.

\textbf{BVP:} Following PMP and for fixed initial states, the equilibrial states $x^*(t)$, actions $u^*(t)$, co-states $\lambda_i^*(t) := \nabla_{x} V_i^*(x^*,t;\theta)$, and values $V^*(x^*,t;\theta)$ for $t\in [0,T]$ are solutions to the following BVP~\cite{pontryagin1966theory}:

{\small
\begin{equation}
\begin{aligned}
    & \dot{x^*} = h(x^*(t), u^*(t)) \\
    & x^*(0) = x_0 \\
    & \dot{\lambda^*}_i = - \nabla_{x} H_i(x^*,u^*, \lambda^*_i(t);\theta) \\
    & \lambda^*_i(T) = - \nabla_{x} c_i(x^*(T);\theta) \\
    & u^*_i(t) = \argmax_{u_i \in \mathcal{U}} H_i(x^*, u_i, \lambda^*_i(t);\theta),\\
    & \dot{V}^*(x^*,t;\theta) = f(x^*,u^*;\theta), \\
    & V^*(x^*,T;\theta) = c(x^*(T);\theta) ~\forall i=1,...,N,
    \label{eq:pmp}
\end{aligned}
\end{equation}}
where $H_i(x,u,\lambda_i,t;\theta) = \lambda_i^T h_i(x,u) - f_i(x,u;\theta)$ is the Hamiltonian for agent $i$. $x_0$ is the initial states. 
% The value is defined as:
% \begin{equation}
%     V_i^*(x_i,x_{-i},t;\theta) = \int_{\tau=t}^T l(x^*,u^*;\theta_i)d\tau + c(x^*(T);\theta_i).
% \end{equation}
Note that $V^*(x,t;\theta)$ is parameterized by all agent parameters due to its implicit dependence on the equilibrial actions $u^*$. We solve Eq.~\eqref{eq:pmp} using a standard BVP solver~\cite{kierzenka2001bvp} with case-specific modifications to be introduced in Sec.~\ref{sec:case}.

\textbf{Value approximation}: Solving the BVPs for given $\theta$ and $x_0$ gives us $V^*$ and $\nabla_x V^*$ for all agents along an equilibrial trajectory starting from $x_0$ and $t=0$. Let this set of values and co-states be $D_v(x_0,\theta)$. We then collect the data $\mathcal{D}_v := \{D_v(x,\theta) ~|~ x \in \mathcal{S}_x, \theta \in \Theta^N\}$ where $\mathcal{S}_x$ is a finite mesh of $\mathcal{X}^N$. This data allows us to build surrogate models for the equilibrial values: $\hat{V}(\cdot,\cdot;\theta,w): \mathcal{X}^N \times [0,T] \rightarrow \mathbb{R}^N$ by solving the following training problem with respect to the surrogate model parameters $w$:
\cutequationup
% {\small
\begin{equation}
\begin{aligned}
    \min_{w} \quad & \sum_{(x,t,V^*,\nabla V^*) \in \mathcal{D}_v} \left(||\hat{V}(x,t;\theta,w)-V^*||^2 \right.\\
    & \left. + C ||\nabla_x \hat{V}(x,t;\theta,w) - \nabla_x V^* ||^2\right).
    \label{eq:valueapp}
\end{aligned}
\end{equation}
% }
Here $C$ balances the matching of values and co-states, and $||\cdot||$ is the $l_2$-norm. To accommodate potential discontinuity in the values, we model $\hat{V}$ using a deep neural network, and derive its co-states through auto-differentiation. Eq.~\eqref{eq:valueapp} can then be solved using a gradient-based solver for all combinations of preferences $\theta \in \Theta^N$. The result is a set of value functions $\mathcal{V} := \{\hat{V}(\cdot,\cdot;\theta,w)\}_{\theta \in \Theta^N}$. Alg.~\ref{alg:app} summarizes the value approximation procedure.

\begin{algorithm}[h]
\SetAlgoLined
\SetKwInOut{Input}{input}
\SetKwInOut{Output}{output}
\Input{$\mathcal{S}_x$, $\Theta^N$, $T$}
\Output{$\{\hat{V}(\cdot,\cdot;\theta,w)\}_{\theta \in \Theta^N}$}
    set $\mathcal{D}_v = \emptyset$, $\mathcal{V} = \emptyset$\;
    \For{each $(x_0, \theta) \in \mathcal{S}_x \times \Theta^N$}{
        solve Eq.~\eqref{eq:pmp} for $D_v(x_0,\theta) = \{x^*(t),\lambda^*(t),V^*(x^*,t;\theta)\}_{t\in[0,T]}$ \;
        $\mathcal{D}_v \leftarrow D_v(x_0,\theta)$\;
    }

    \For{each $\theta \in \Theta^N$}{
        solve Eq.~\eqref{eq:valueapp} for $\hat{V}(\cdot,\cdot;\theta,w)$\;
        $\mathcal{V} \leftarrow \hat{V}(\cdot,\cdot;\theta,w)$\;
    }

\caption{Value approximation}
\label{alg:app}
\end{algorithm}
\vspace{-0.1in}
\textbf{Action-value approximation:}
We approximate the action-value at time $t$ using the Hamiltonian. 
% by
% \begin{equation}
% \begin{aligned}
%     Q((x,k\Delta t),u;\theta) = & \hat{V}(x,k\Delta t;\theta,w) + \\ 
%     &\nabla_x \hat{V}(x,k\Delta t;\theta,w)^T h(x,u) \Delta t, \\
% \end{aligned}
% \end{equation}
% where $\Delta t$ is the step size. 
Note that we need to consider time as part of the state since the game has a finite time horizon.

\cutsectionup
\section{Case study}
\cutsectiondown
\label{sec:case}
The goal of the case study is to identify interaction settings where empathetic agents together perform ``better'' than non-empathetic ones. In order to perform a thorough study and due to space limitation, we focus on an uncontrolled intersection case and discuss experimental settings, hypotheses and  analyses as follows.

\subsection{Uncontrolled intersection}\label{subsec:uncontrol_insc}
\begin{figure}
    \centering
    \includegraphics[width=\linewidth]{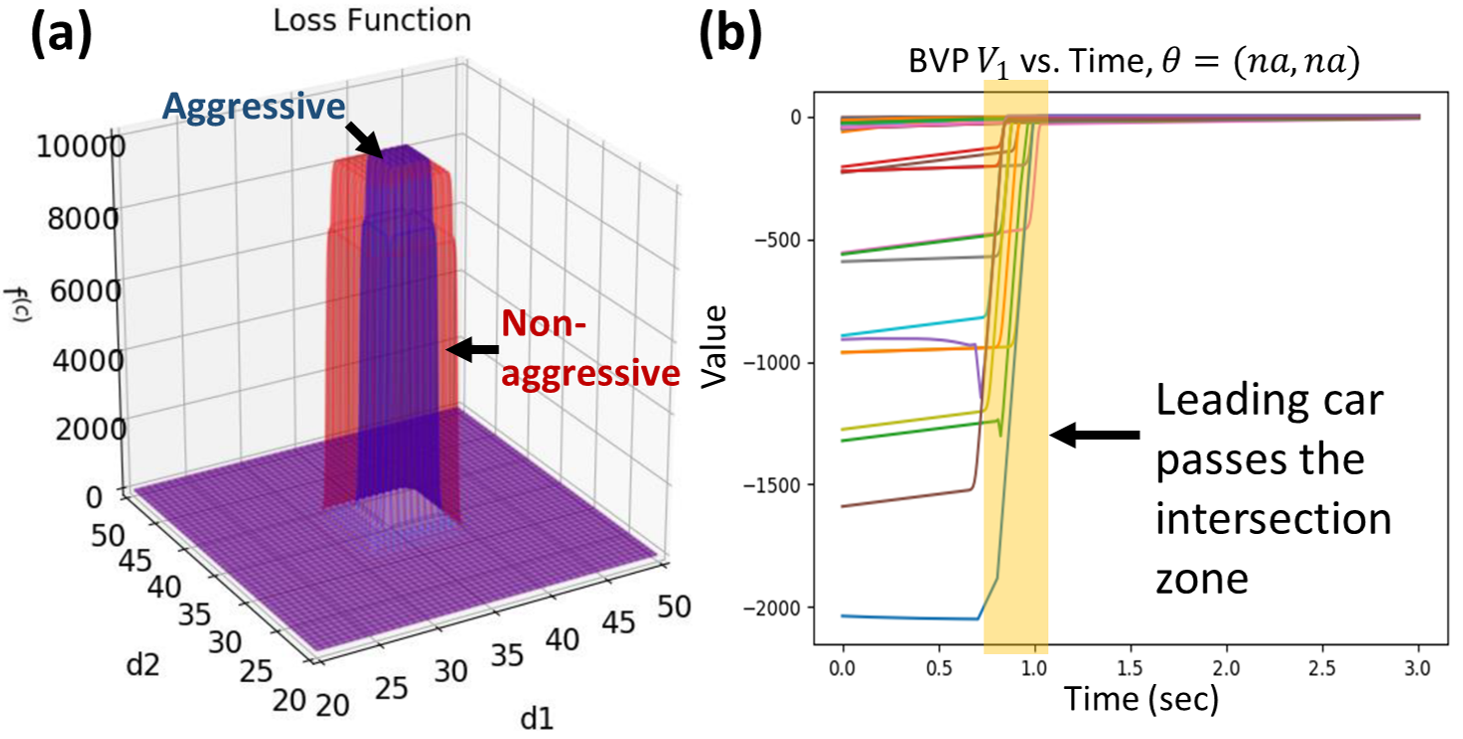}
    \vspace{-0.25in}
    \caption{(a) Collision loss in $(d_1,d_2)$ (b) Equilibrial value of one agent along time when both agents are non-aggressive}
    \label{fig:setting}
    \vspace{-0.25in}
\end{figure}

This case models the interaction between two cars at an uncontrolled intersection specified in Fig.~\ref{fig:summary}a. The state of agent $i$ is defined by the agent's position $d_i$ and its velocity $v_i$: $x_i=(d_i,v_i)$. The individual state space is set as $\mathcal{X}= [15, 20]m \times \{18, 18\}m/s$, where the initial velocity is fixed for visualization purpose and can be extended in future work. The action of agent $i$ is defined as its acceleration rate, and the action space as $\mathcal{U} = [-5,10]m/s^2$. The instantaneous reward function is
\begin{equation}
    f_i(x,u;\theta) = f^{(e)}(u_i) + f^{(c)}(x;\theta_i),
\end{equation}
% $f_i(x,u;\theta) = f^{(effort)}(u_i) + f^{(collision)}(x;\theta_i)$. 
where $f^{(e)}(u_i) = -u_i^2$ is a negative effort loss and
\begin{equation}
    f^{(c)}(x;\theta_i) = -b\prod_{{i,j}=\{(1,2),(2,1)\}} \sigma_1(d_i,\theta_i)\sigma_2(d_j)
\end{equation}
% $f^{(collision)}(x;\theta_i) = -\prod_{i=\{1,2\},j=\{2,1\}} \beta\sigma(x_i-R_{i}/2 + L_{i}/2 + \theta_j W_{j}/2)\sigma(-x_i + R_{i}/2 + L_{i}/2 + W_{j}/2)$ 
models a negative penalty for collision. Here 
{\small
\begin{equation}
    \sigma_1(d,\theta) = \left(1+\exp(-\gamma (d-R/2+\theta W/2))\right)^{-1};
\end{equation}
\begin{equation}
    \sigma_2(d) = \left(1+\exp(\gamma (d-R/2-W/2-L))\right)^{-1};
\end{equation}
}
$b=10^4$ sets a high loss for collision; $\gamma=10$ is a shape parameter; $R$, $L$, and $W$ are the road length, car length, and car width, respectively (Fig.~\ref{fig:summary}a). $\theta_j$ denotes the aggressiveness (sensitivity to collision) of the agent. Fig.~\ref{fig:setting}a visualizes $f^{(c)}$ along $d_1$ and $d_2$. The terminal loss is defined as $c_i(x) = \alpha d_i(T) - (v_{i}(T)-v_{0})^2$, where $\alpha = 10^{-6}$, i.e., the agent is rewarded for moving forward and restoring its initial speed at $T$. We adopt a simple dynamical model:  
\begin{equation}
    {\dot x_{i}(t)}=\left[
\begin{array}{ccc}
    0\hspace{0.4cm} 1 \\
    0\hspace{0.4cm} 0\\
\end{array}
\right]
\left[
\begin{array}{ccc}
    d_{i}(t) \\
    v_{i}(t) \\
\end{array}
\right]
+
\left[
\begin{array}{ccc}
    0 \\
    1 \\
\end{array}
\right]
u_{i}(t).
\end{equation}
We set $\Theta = \{1,5\}$ and $\Lambda = \{0.1, 0.5\}$ as common knowledge. Note that $\theta_i=1$ ($\theta_i=5$) represents an aggressive (non-aggressive) agent; $\lambda_i=0.1$ ($\lambda_i=0.5$) represents a noisy (less-noisy) decision model. We solve BVPs on $\mathcal{S}_x$, which is a meshgrid of $\mathcal{X}$ with an interval of $0.5m$ for both $d_1$ and $d_2$.

\begin{figure}
    \centering
    \includegraphics[width=0.95\linewidth]{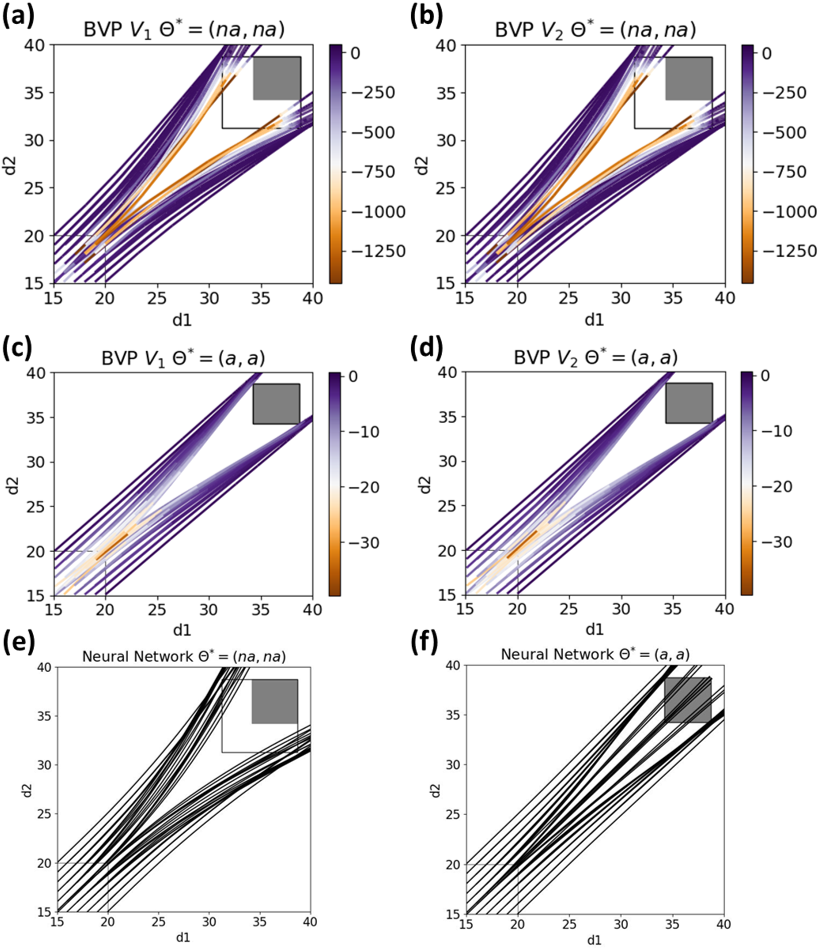}
    \vspace{-0.1in}
    \caption{Interactions b/w non-aggressive (a,b) and aggressive agents (marked as (na,na) and (a,a) respectively), colored by the equilibrial values of agent 1 (a,c) and 2 (b,d).  Negative values due to collision penalty. Smaller values in (a,a) due to low sensitivity to close distances by aggressive agents. (e,f) Interactions reproduced through value approximation for (na,na) and (a,a).}
    \label{fig:bvp}
    \vspace{-0.1in}
\end{figure}

\textbf{Solving BVPs}: BVP solutions for complete-information differential games are known to be dependent on the initial guess of state and co-state trajectories~\cite{johnson2009numerical}. Specific to our case, it can be shown from Eq.~\eqref{eq:pmp} that collision avoiding behavior can only be derived when numerical integration over $\partial f^{(c)}/\partial d_i$ can be correctly performed. This integration, however, is challenging since $\partial f^{(c)}/\partial d_i$ resembles a mixture of delta functions, and therefore requires dense sampling in the space of $(d_1,d_2)$ where the collision happens. To this end, we predict two time stamps, $t_1$ and $t_2$, respectively corresponding to (1) when the second car enters and (2) when the first car leaves the intersection zone. The prediction is done by assuming that the leading car moves at its initial velocity and the trailing car takes maximum deceleration. We then densely sample around $t_1$ using $\{t_1 \pm 1.25\times 10^{-6}k\}_{k=0}^{800}$. These time stamps along with the approximated agent actions provide an initial guess for the system states and co-states. Fig.~\ref{fig:bvp}a-d demonstrates equilibrial trajectories in the space of $(d_1,d_2)$ when both agents are non-aggressive and aggressive.

\textbf{Approximating values}: We notice that there exist abrupt changes in the value along time and space in the BVP solutions, due to the high penalty of collision and close calls (Fig.~\ref{fig:setting}b), i.e., after the two agents pass each other, which in some cases incurs high loss due to close distances between the two, the value increases significantly. We found that conventional fully-connected network architectures cannot effective learn this unique structure, and therefore propose the following value network architecture:
\cutequationup
% {\small
\begin{equation}
    \hat{V}(x,t;\theta,w) = \eta f_1(x,t;\theta,w) + (1-\eta) f_2(x,t;\theta,w),
\cutequationdown
\end{equation}
% }
where $f_1$ and $f_2$ follow the same architecture: \texttt{fc5-(fc16-tanh)$\times$3-(fc2-tanh)}, where \texttt{fc}$n$ represents a fully connected layer with $n$ nodes and \texttt{tanh} is the hyperbolic tangent activation. $\eta$ is a sigmoid function that determines whether one of the agents have passed the intersection zone. Training data are collected from $|\mathcal{S}_x|=121$ BVP solutions, and test data from another 36 solutions samplmed in $\mathcal{X}^2$. We use ADAM~\cite{kingma2014adam} with learning rate 0.01. We set the hyperparameter $C=1$.  Fig.~\ref{fig:bvp}e,f illustrate the approximated trajectories in $(d_1,d_2)$ considering complete information, where actions are chosen by maximizing the approximated action-values using the resultant value networks, when both agents are respectively non-aggressive and aggressive. Some remarks: While value approximation is not perfect (test relative MAE of $15.64\%$ ($12.17\%$) for non-aggressive (aggressive) cases), the approximated equilibria are mostly acceptable. We do face an intrinsic challenge in learning the values when both agents have the same initial states and are aggressive, potentially due to a combination of relatively high error in co-state approximation ($83.73\%$ relative MAE) and the nonuniqueness of equilibria in these scenarios, i.e., either of the agent can yield or move first.

\cutsubsectionup
\subsection{Driving scenarios}
An incomplete-information driving scenario is a tuple $s = <x_0,p_0(\beta),\theta^{*},l>$ specifying the initial state, prior belief, true parameters, and estimation types. 
% We denote by $\mathcal{S}:=\mathcal{X}_0 \times \mathcal{P}_0 \times \Theta \times \mathcal{L}$ the set of driving scenarios. 
% \textbf{The initial state set $\mathcal{X}_0$:} 
We pick initial states from $\mathcal{S}_x$, and parameters (aggressiveness) from $\Theta$. We use $a$ for aggressive and $na$ for non-aggressive agents, and $n$ for noisy and $ln$ for less-noisy agents during belief updates.
% \begin{table}[]
%     \centering
%     \begin{tabular}{c|c|c|c|c}
%     \hline
%         case & $\theta^{(a)}$ & $\theta^{(na)}$ & $\lambda^{(n)}$ & $\lambda^{(nn)}$ \\
%         \hline
%         \texttt{intersection} & *** & *** & *** & *** \\
%         \texttt{lane changing} & *** & *** & *** & *** \\
%         \hline
%     \end{tabular}
%     \caption{Agent parameter levels}
%     \label{tab:parameter}
% \end{table}
\textbf{The prior common belief set $\mathcal{P}_0$:} With the above parameter settings, each element of the prior belief set $\mathcal{P}_0$ is a 4-by-4 matrix containing joint probabilities for all 16 agent parameter combinations. Each dimension of the matrix follows the order $(a,n), (a,ln), (na,n), (na,ln)$, e.g., the 1st row and 2nd column of the matrix represents $\Pr\left(\beta_1 = (a,n), \beta_2 = (a,ln)\right)$. To constrain the scope of the studies, we set to prior believe that agents are mostly rational ($\Pr(\lambda_{1,2} = ln) = 0.8$), and explore two cases on $\theta$: (1) Everyone believes that everyone is most likely non-aggressive ($\Pr(\theta_{1,2} = na) = 0.8$) or most likely aggressive ($\Pr(\theta_{1,2} = a) = 0.8$). This reduces $\mathcal{P}_0$ to $\{p_0^{na}, p_0^{a}\}$, where
% listed in Tab.~\ref{tab:belief}
% \begin{table}[]
%     \centering
%     \begin{tabular}{c|c|c|c|c|c|c|c|c|c|}
%     \hline
%         \multicolumn{5}{c}{\texttt{intersection}} & \multicolumn{5}{c}{\texttt{lane changing}}\\
%         \hline
%         $p(\beta)$ & a,n & a,nn & na,n & na,nn & $p(\beta)$ & a,n & a,nn & na,n & na,nn & \\
%         a,n &  
%     \end{tabular}
%     \caption{Caption}
%     \label{tab:my_label}
% \end{table}
% Use this for the journal version
% \begin{equation}
% \begin{aligned}
%     & p_0^{na} = 10^{-3}\begin{bmatrix}
%     1.6 & 6.4 & 6.4 & 25.6 \\
%     6.4 & 25.6 & 25.6 & 102.4 \\
%     6.4 & 25.6 & 25.6 & 102.4 \\
%     25.6 & 102.4 & 102.4 & 409.6 
%     \end{bmatrix},\\
%     & p_0^{a} = 10^{-3}\begin{bmatrix}
%      6.4 & 25.6 & 1.6 & 6.4 \\
%     25.6 & 102.4 & 6.4 & 25.6 \\
%     25.6 & 102.4 & 6.4 & 25.6 \\
%     102.4 & 409.6 & 25.6 & 102.4 
%     \end{bmatrix},
% \end{aligned}
% \end{equation}
% where 
$p_0^{na}$ ($p_0^{a}$) is the common prior where everyone is believed to be non-aggressive (aggressive).
% \textbf{Belief update frequency:} We set $\Delta\mathcal{T}=\{10ms, 250ms, 1s\}$. Note that $250ms$ is the average human reaction time to visual stimulus. Therefore $\Delta\mathcal{T}$ covers super-, average-, and sub-human update frequencies.
\textbf{Parameter estimation type:} We set
$\mathcal{L}=\{(e,e),(ne,ne)\}$ where $e$ stands for ``empathetic'' and $ne$ for ``non-empathetic''. Using Alg.~\ref{alg:sim} and by setting a time interval of $0.05s$, interaction trajectories can be computed for each driving scenario $s$. The resultant values at $t=0$ are denoted by $v(s)$.
\textbf{Evaluation metrics}: Lastly, we measure the goodness of empathetic and non-empathetic estimations using social value, i.e., the sum of the individual values at $t=0$: $\bar{v}(s) = \sum_{i=1}^N v_i(s)$. \textbf{Videos}: See supplementary video for animated interactions.
% \begin{enumerate}
% \item \textbf{Social} This metric aims to reflect on the social orientation of the agent. We look into the amount of courtesy each model presents by comparing ratio $\frac{f_i(x,u,\theta}{f_{-i}(x,u,\theta)}$.
% \item \textbf{Behavioral} This metric quantify the efficiency of each model. We compare the sum of the value function $\sum_{i=1}^N v_i(s)$ for both models. 
% \item \textbf{Intent} This metric aims to analyze each model's inference quality. We look into the accuracy of a-posteriori inference of each agent. For a fixed $\theta$ and $\beta$, for a given iteration, we verify the maximum a-posteriori $\theta^{MAP}$ with the original $\theta$.  
% \end{enumerate}
% For comparison, we introduce \textit{social value} as the sum of agent values at $t=0$: $\bar{v}(s) = \sum_{i=1}^N v_i(s)$.
\cutsubsectionup
\subsection{Hypotheses} 
The following hypotheses concerning two driving scenarios $s^{(1)}$ and $s^{(2)}$ will be tested empirically:
\begin{enumerate}
    \item \textit{Empathy leads to higher social value when agents are unknowingly aggressive (or non-aggressive):} Let $l^{(1)} = (e,e)$, $l^{(2)} = (ne,ne)$, $\theta^{*(1)} = \theta^{*(2)} = (a,a)$ (or $(na,na)$), $p_0^{(1)} = p_0^{(2)} = p_0^{na}$ (or $p_0^{a}$). There exists $\mathcal{X}_0' \subset \mathcal{X}_0'$, such that for all $x_0 \in \mathcal{X}_0'$,  $\bar{v}(s^{(1)}) > \bar{v}(s^{(2)})$.
    
    \item \textit{Empathy leads to higher social value when agents are knowingly aggressive (or non-aggressive):} The same as Hypothesis 1, except that the common beliefs are set to be consistent with the truth parameters.
    % \item \textit{Empathy is more useful when belief update frequency is low:} Use the same settings as in 1) and 2). Let $\bar{v}_{\Delta t, x_0}(s)$ be the social value for specific $\Delta t$ and $x_0$. Introduce $\Delta \bar{v}_{\Delta t} = \frac{1}{|\mathcal{X}_0|} \sum_{x_0 \in \mathcal{X}_0} \bar{v}_{\Delta t, x_0}(s^{(1)}) - \bar{v}_{\Delta t, x_0}(s^{(2)})$, i.e., $\Delta \bar{v}_{\Delta t}$ is the expected social value advantage of empathetic parameter estimation for a given belief update frequency. If $\Delta t_1 > \Delta t_2$, then $\Delta \bar{v}_{\Delta t_1} > \Delta \bar{v}_{\Delta t_2}$.
\end{enumerate}
\cutsubsectionup
\subsection{Results and analysis}
\begin{figure}
    \centering
    \includegraphics[width=0.90\linewidth]{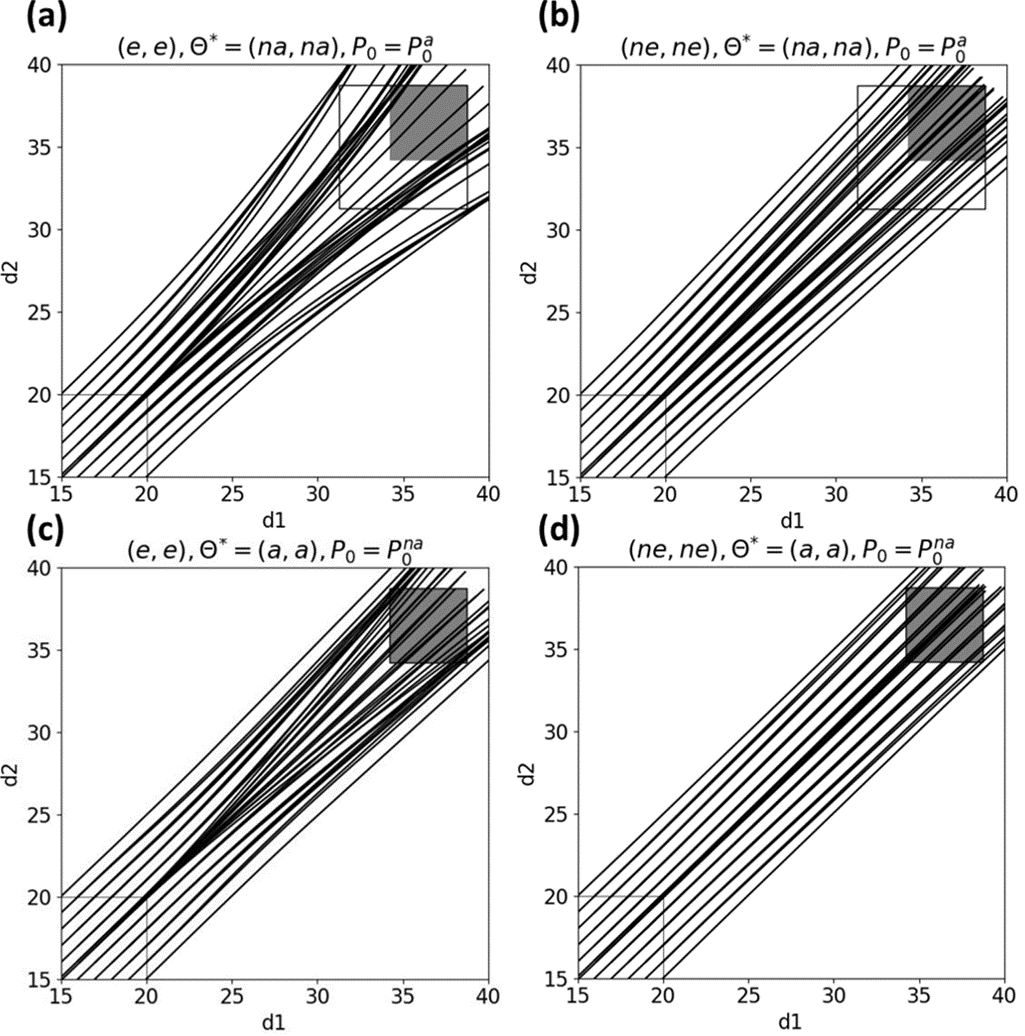}
    \vspace{-0.2in}
    \caption{
    % $\bar{v}(s^{(1)}) - \bar{v}(s^{(2)})$ on $\mathcal{S}_x$ 
    Interactions when common belief mismatches with true reward parameters. (a,b) Unknowingly non-aggressive, (c,d) Unknowingly aggressive. (a,c) Empathetic, (b,d) Non-empathetic.}
    \label{fig:inconsistent}
    \vspace{-0.25in}
\end{figure}

\begin{figure}[t]
    \centering
    \includegraphics[width=0.9\linewidth]{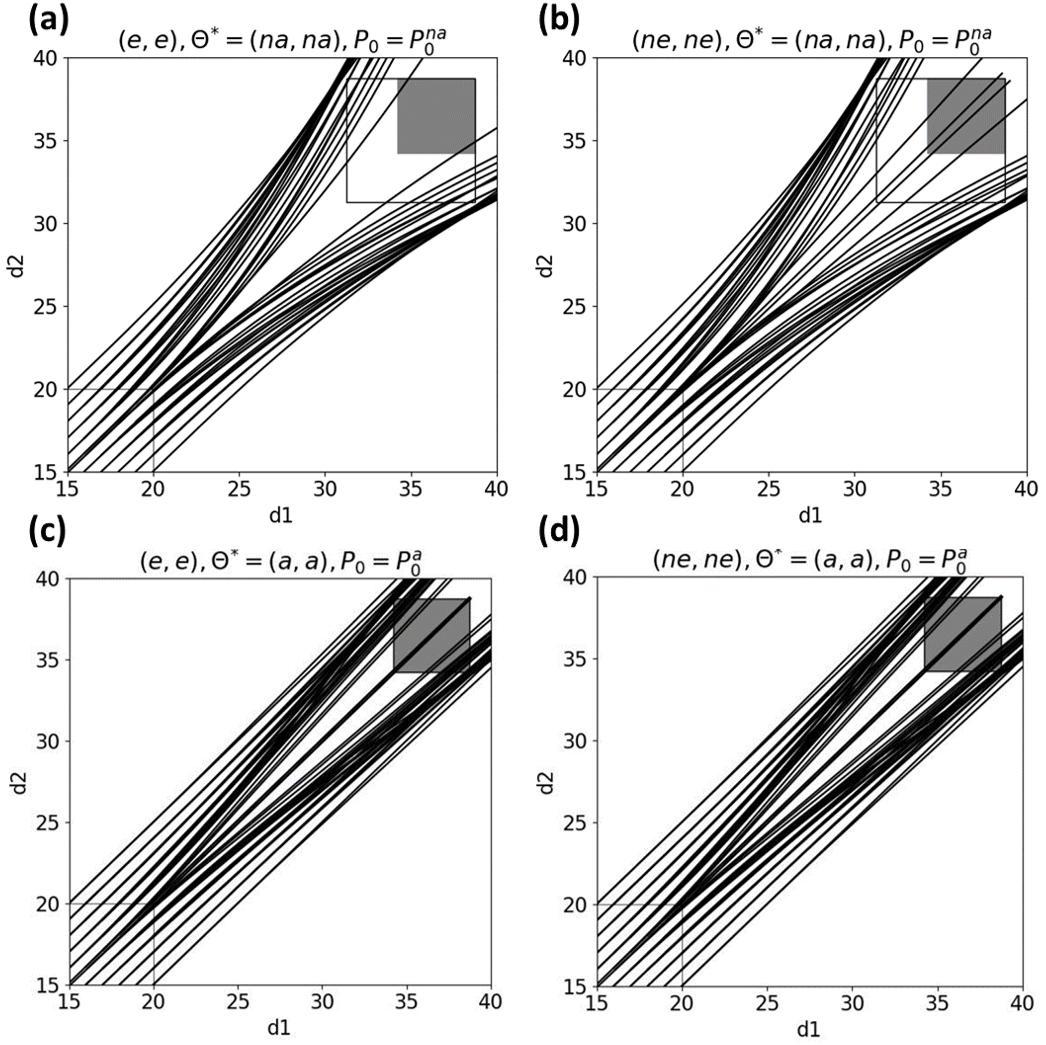}
    \vspace{-0.1in}
    \caption{
    % $\bar{v}(s^{(1)}) - \bar{v}(s^{(2)})$ on $\mathcal{S}_x$ 
    Interactions when common belief matches with true reward parameters. (a,b) Knowingly non-aggressive, (c,d) Knowingly aggressive. (a,c) Empathetic, (b,d) Non-empathetic.}
    \label{fig:consistent}
    \vspace{-0.1in}
\end{figure}

\begin{figure}
    \centering
    \includegraphics[width=0.9\linewidth]{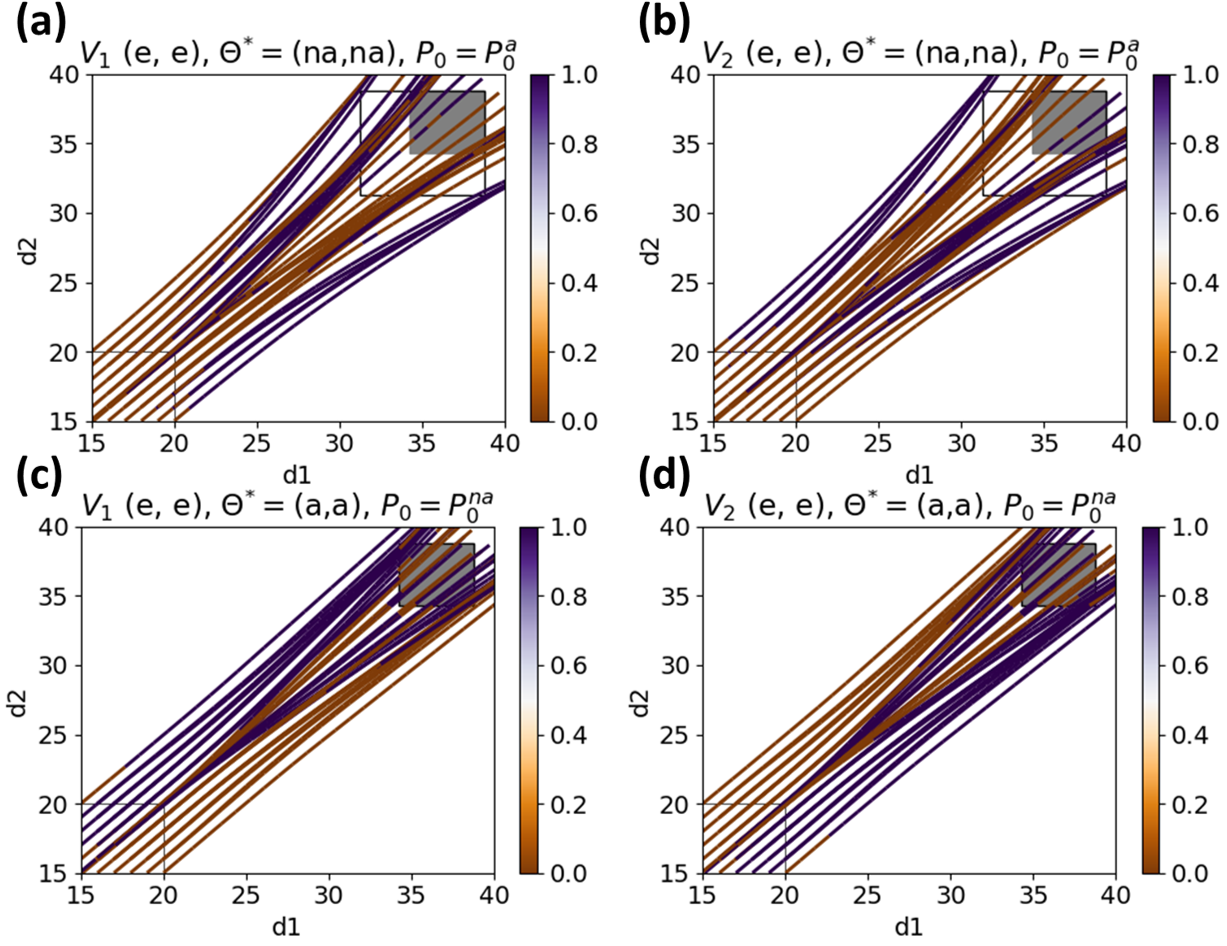}
    \vspace{-0.15in}
    \caption{
    % $\bar{v}(s^{(1)}) - \bar{v}(s^{(2)})$ on $\mathcal{S}_x$ 
    Color-coding of the policy choices by empathetic agents, for non-aggressive (a,b) and aggressive (c,d) scenarios}
    \label{fig:policye}
    \vspace{-0.1in}
\end{figure}

\begin{figure}[h!]
    \centering
    \includegraphics[width=0.9\linewidth]{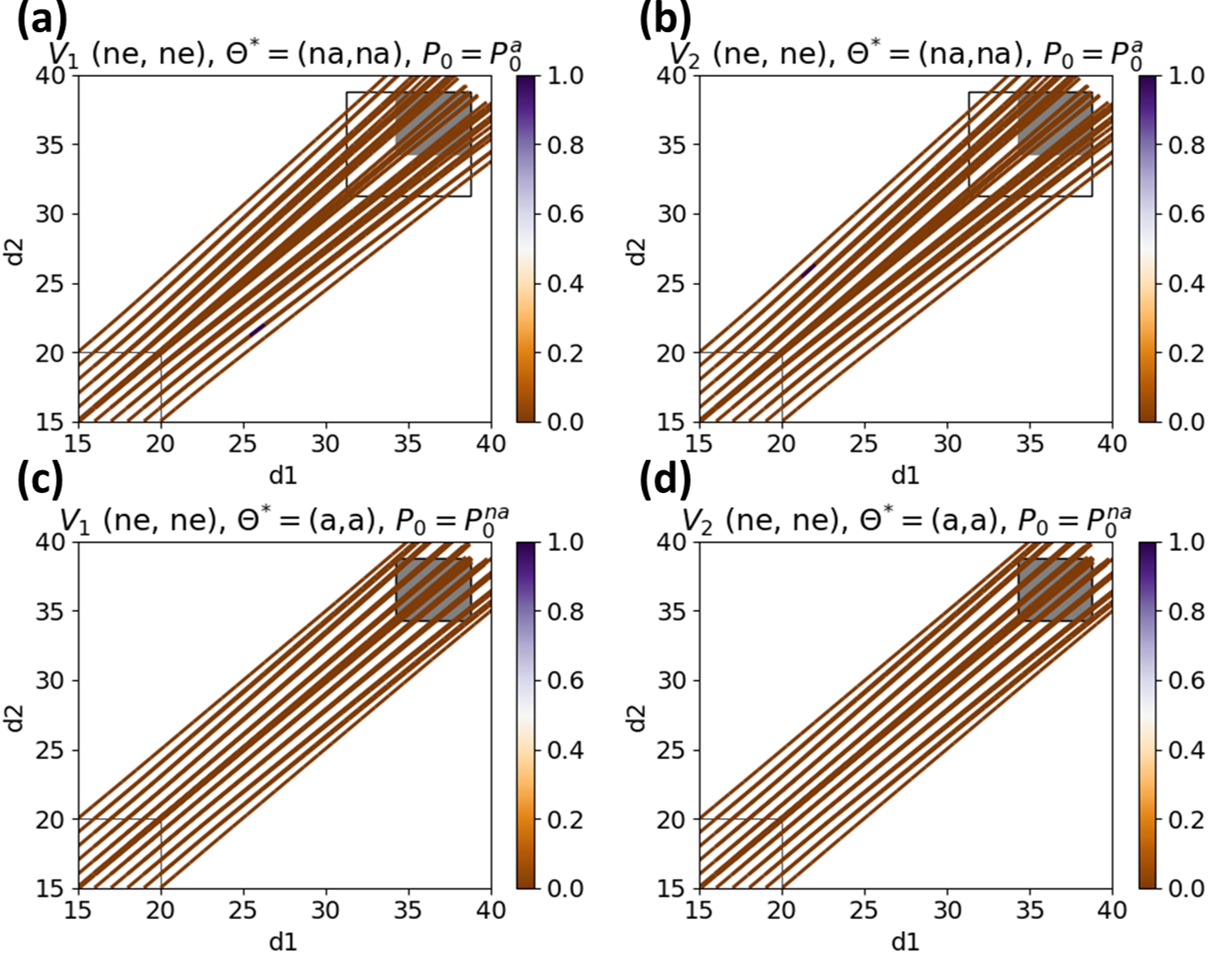}
    \vspace{-0.10in}
    \caption{
    % $\bar{v}(s^{(1)}) - \bar{v}(s^{(2)})$ on $\mathcal{S}_x$ 
    Color-coding of the policy choices by non-empathetic agents, for non-aggressive (a,b) and aggressive (c,d) scenarios}
    \label{fig:policyne}
    \vspace{-0.1in}
\end{figure}

\cutsubsectiondown
Hypothesis 1 passes the test (Fig.~\ref{fig:inconsistent}), suggesting that being empathetic leads to higher social values (i.e., less collisions in the intersection case) when common belief is wrong. Hypothesis 2 passes the test (Fig.~\ref{fig:consistent}), although results suggest that when the common belief is consistent with the true parameters, empathy does not play a significant role. Also notice that matching between belief and parameters help improve the interactions.
To understand how empathy helps, we inspect whether agents choose the correct policies (among (na,na), (na,a), (a,na), and (a,a)) at each time step during the interaction following Alg.~\ref{alg:sim}. Specifically, when agents are non-aggressive, the correct policy follows $\hat{\beta}=(na,na)$, vice versa. In Figs.~\ref{fig:policye} and \ref{fig:policyne}, we color-code the correct (1) and incorrect (0) choices of policies for both agents. Results show that empathetic agents tend to choose the correct policies when they are trailing. We conjecture that this is due to the fact that the actions of the leading agent are intrinsically more effective at signaling, i.e., its lower acceleration suggests that it does not care much about potential close distances and thus its high aggressiveness. On the other hand, non-empathetic agents never choose the correct policies.

% \begin{figure}[t]
%     \centering
%     \includegraphics[scale=0.5]{images/EVNANA.png}
%     \caption{Empathetic with Non Aggressive Baseline Comparison}
%     \label{fig:emp_v_nana}
% \end{figure}
% \begin{figure}[t]
%     \centering
%     \includegraphics[scale=0.5]{images/EVNE_NANA.png}
%     \caption{Empathetic with Non-Empathetic Non Aggressive Comparison}
%     \label{fig:emp_v_nonemp}
% \end{figure}
% \begin{figure}[t]
%     \centering
%     \includegraphics[scale=0.4]{images/EVAA.png}
%     \caption{Empathetic with Aggressive Baseline Comparison}
%     \label{fig:emp_v_aa}
% \end{figure}
% \begin{figure}[t]
%     \centering
%     \includegraphics[scale=0.5]{images/EVNE_AA.png}
%     \caption{Empathetic with Non-Empathetic Aggressive Comparison}
%     \label{fig:emp_v_nonemp}
% \end{figure}

% \cutsectionup


\section{Conclusions}
% \cutsectiondown
\label{sec:conclusion}
Using an uncontrolled intersection case, we studied the utility of empathetic parameter estimation in a two-agent incomplete-information dynamic game. We showed that empathy helped agents choose policies that led to higher social values when agents had common beliefs inconsistent with their true parameters. When belief and parameters were consistent, empathy did not hurt social values. While its findings should be tested under a larger set of driving scenarios (e.g., roundabout and lane changing), this study provides a methodology for systematically evaluating the utility of empathy in incomplete-information dynamic games.
% We complete with a few directions for future investigation: (1) To our best knowledge, the equilibrial value approximation technique explored in this paper is among the first for general-sum differential games. Generalizing the value network to a full set of common driving scenarios will be useful for other game-theoretic studies on modeling human driver behaviors. (2)  ***

\newpage
\bibliography{ref}
\bibliographystyle{IEEEtran}
\end{document}